\documentclass[sigconf]{acmart}
\usepackage{booktabs} 

\usepackage{amssymb}
\usepackage{amsthm, amsmath}
\usepackage{algorithmic}
\usepackage[ruled]{algorithm2e}
\usepackage{color}
\usepackage{url}
\usepackage{xfrac}
 
\usepackage{subfigure}
\usepackage{tablefootnote}

\usepackage{multirow}
\usepackage{tabularx}
\usepackage{multirow}
\usepackage{balance}

\begin{document}
\title{Multiresolution Graph Attention Networks \\for Relevance Matching}

\fancyhead{}
\settopmatter{printfolios=false}


\author{Ting Zhang$^1$, Bang Liu$^1$, Di Niu$^1$, Kunfeng Lai$^2$, Yu Xu$^2$}
       \affiliation{$^1$University of Alberta, Edmonton, AB, Canada}
 \affiliation{$^2$Mobile Internet Group, Tencent, Shenzhen, China}

\begin{abstract}
A large number of deep learning models have been proposed for the text matching problem, which is at the core of various typical natural language processing (NLP) tasks.
However, existing deep models are mainly designed for the semantic matching between a pair of short texts, such as paraphrase identification and question answering, and do not perform well on the task of relevance matching between \textit{short-long} text pairs.
This is partially due to the fact that the essential characteristics of short-long text matching have not been well considered in these deep models. 
More specifically, these methods fail to handle extreme length discrepancy between text pieces and neither can they fully characterize the underlying structural information in long text documents.

In this paper, we are especially interested in relevance matching between a piece of short text and a long document, which is critical to problems like query-document matching in information retrieval and web searching. 
To extract the structural information of documents, an undirected graph is constructed, with each vertex representing a keyword and the weight of an edge indicating the degree of interaction between keywords. 
Based on the keyword graph, we further propose a \textit{Multiresolution Graph Attention Network} to learn multi-layered representations of vertices through a Graph Convolutional Network (GCN), and then match the short text snippet with the graphical representation of the document with the attention mechanisms applied over each layer of the GCN.
Experimental results on two datasets demonstrate that our graph approach outperforms other state-of-the-art deep matching models.
\end{abstract}

\copyrightyear{2018}
\acmYear{2018}
\setcopyright{acmcopyright}
\acmConference[CIKM '18]{The 27th ACM International Conference on
Information and Knowledge Management}{October 22--26, 2018}{Torino,
Italy}
\acmBooktitle{The 27th ACM International Conference on Information and
Knowledge Management (CIKM '18), October 22--26, 2018, Torino, Italy}
\acmPrice{15.00}
\acmDOI{10.1145/3269206.3271806}
\acmISBN{978-1-4503-6014-2/18/10}

%



\maketitle
\section{Introduction}
\label{sec:introduction}

Matching two pieces of text has long been a core research problem underlying numerous natural language processing tasks.
The past few years have seen the great success of deep models ~\cite{hu2014convolutional,qiu2015convolutional,wan2016deep,pang2016text} for semantic matching tasks such as question answering (QA)~\cite{wang2017bilateral}, paraphrase identification~\cite{yin2015convolutional} and automatic conversation~\cite{ji2014information} etc.
However, it is still challenging to estimate the relevance between a pair of \textit{short} and \textit{long} text pieces. 
For example, in query-document matching, user queries usually contain a few words, while the lengths of documents could vary from hundreds to thousands of words. Given rich semantic and syntactic structures that exist in long documents and the extreme discrepancy between the lengths of queries and documents, accurately estimating the relevance between them is hard.

Existing methods for text matching are typically categorized into three types including unsupervised metrics \cite{kusner2015word}, feature-based models and deep matching models \cite{hu2014convolutional,qiu2015convolutional,wan2016deep,pang2016text}. 
For unsupervised metrics, documents are transferred to vectors with representation methods such as bag-of-words (BOW). 
Then the distance between vectors are calculated according to metrics like euclidean distance, cosine similarity and so on. 
However, such approaches are principally based on the term frequency and ignore the semantic structures of natural language. 
Thus leading to poor performance for complicated tasks. 
Feature-based models, or feature engineering \cite{wang2017irgan} rely on hundreds or thousands of handcrafted features. 
In reality, search engines also depend on other auxiliary information like click history, ad hoc rules and metadata, etc., to boost the query-document matching performance. Obviously, handcrafting features is time-consuming, possibly incomplete and application-specific.

Recently, lots of deep models have also been applied to text matching, e.g., \cite{hu2014convolutional,qiu2015convolutional,wan2016deep,pang2016text}, which can be divided into two categories depending on the model structures: representation-focused and interaction-focused.
Representation-focused models \cite{qiu2015convolutional,wan2016deep} take the word embedding sequences of a pair of text objects as the inputs, and learn their  intermediate contextual representations through Siamese neural networks, on which final scoring is performed. 
While interaction-focused deep models \cite{hu2014convolutional,pang2016text} focus on local interactions between two pieces of text and learn the complex interaction patterns with deep neural networks. 
Comparing to other methods, deep matching models are generalized while maintaining high accuracy in various NLP tasks.

However, we show that most existing deep models can not yield satisfactory performance for relevance matching between a pair of \textit{short} and \textit{long} text objects. It is partially due to the essential differences between semantic matching and relevance matching.
Semantic matching tasks, such as paraphrase identification, concentrate on identifying the semantic meaning and inferring the semantic relations between two pieces of text. While relevance matching tasks, such as query document matching in information retrieval, care more about whether the query and document are related or not instead of whether they express the same semantic meaning or not.
We figured out that most existing deep matching models \cite{hu2014convolutional,qiu2015convolutional,wan2016deep,pang2016text} mainly concern semantic matching rather than relevance matching.
Also, we point out that current deep models \cite{hu2014convolutional,qiu2015convolutional,wan2016deep,pang2016text} are effectively dealing with text snippets, e.g., a pair of sentences, but have difficulty handling extreme short text and long documents. On one hand, encoding the query consisting of only few words with complicated deep models usually results in excessive deformation.
On the other hand, it is more likely to introduce ``noise'' and redundant information when dealing with long documents using deep models.

To address the above problems, we propose a deep relevance matching model based on graph and attention mechanisms to improve the matching between a pair of short and long text objects. 
We show that an appropriate semantic representation, beyond a linear sequence of word vectors~\cite{pennington2014glove}, of a document plays a central role in relevance matching. 
Documents are represented as undirected, weighted \textit{Keyword Graph}, in which each vertex is a keyword in the document, and the weight of each edge indicates the relevance degree between two corresponding vertices. Such a graphical representation helps to reveal the inner structures of a document. Based on such representation, the problem of relevance matching is transformed into a query-graph matching problem. 

To match the query and document graph, we designed a novel deep matching model, namely \textit{Multiresolution Graph Attention Network} (MGAN). It learns multiresolution representations for each vertex through a multi-layered graph convolutional networks (GCN), an emerging variant of convolutional neural networks that specifically encodes graphs. 
Moreover, we develop deeper insights into the GCN ~\cite{kipf2016semi} and improve it to better cope with weighted graphs.
By applying attention mechanisms to word vectors of the query with the keyword representations learned by each layer of the GCN, MGAN is able to characterize the relevance between the query and keywords of the document, utilizing multiresolution representations of keywords generated in different layers.
To handle the varying number of keywords in different documents, a \textit{rank-and-pooling} strategy is proposed to sort and select keyword vertices. In each layer, we choose a fixed number of query-keyword matching results, and concatenate them together. 
The final relevance score is generated by feeding the concatenated matching vector into a multilayer perceptron network.

We evaluated our model on the Ohsumed dataset and the NFCorpus dataset. 
Experimental results demonstrate that our model boasts significantly improved performance compared with existing state-of-the-art deep matching models. 

The remainder of this paper is organized as follows.
Sec.~\ref{sec:problem} formally introduces the problem of relevance matching as well as its characteristics.
Sec.~\ref{sec:graph} presents the keyword graph construction of long documents.
In Sec.~\ref{sec:match}, we propose the Multiresolution Graph Attention Network for relevance matching of short-long text pairs. 
Experimental results are demonstrated in Sec.~\ref{sec:eval}.
We review the related literature in Sec.~\ref{sec:related} and finally conclude the paper in Sec.~\ref{sec:conclude}.

\section{Relevance Matching}
\label{sec:problem}

In this section, we formally introduce the problem of relevance matching, and show the differences between relevance matching and semantic matching. Most importantly this section serves to point out the challenges in matching the relevance between a piece of short text and a long document, such as the query and document matching.

Denote a query as $q$ and a text document as $d$. Given a query-document pair $(q,d)$, the relevance matching problem can be formalized as:
\begin{equation}
\label{eq:qd-matching}
r = \mathcal{F}(\phi_q(q), \phi_d(d))
\end{equation}
where $\phi_q$ and $\phi_d$ are representation functions that map query and document to their feature space. $\mathcal{F}$ is the scoring function based on the interactions between query and document. The relevance score $r$ can be binary or numerical: binary $r$ indicates whether the text pair is related or not, while numerical $r$ reflects the relevance degree between a query and a document. 

A lot of deep matching models have been proposed \cite{qiu2015convolutional,wan2016deep,hu2014convolutional,pang2016text}, and most of them have only been demonstrated to be effective on a set of NLP tasks such as semantic textual similarity, paraphrase identification, question answering \cite{guo2016deep} and so on. However, when these deep models are applied on relevance matching problem in Eq.~\ref{eq:qd-matching} such as the task of query document matching, their performance is usually disappointing.

This is due to some fundamental differences between the tasks of semantic matching and relevance matching, as pointed out by \cite{guo2016deep}.
The goal of semantic matching is to understand the semantic meaning of the text or infer the relationship between two pieces of text, which are usually homogeneous sentences. 
However, relevance matching focuses on deciding whether two pieces of text describe the relevant topics. 
For example, ``A man is playing basketball.'' is semantically similar with ``A man is playing football.'', but these two sentences are not relevant.
Another example is that ``Tom is chasing Jerry in the yard.'' is relevant to ``Tom is chased by Jerry in the yard.'', but they are not semantically similar. 
In the semantic matching, since sentences usually consist of different grammatical structures, it is more beneficial to implement syntactic analysis.
For relevance matching, it emphasizes more on the term matching signals between the query and document.
Actually, most existing models are concerned about \textit{semantic matching} tasks, such as paraphrase identification, question answering \cite{guo2016deep} and so on, but few of them consider the characteristics of the relevance matching.

Besides, in the task of query document matching, query and document vary considerably in text length and provide unbalanced information for directly matching. 
The query is usually extremely short and consists of only few words, while the length of document varies from tens of words to tens of thousands of words.  
Current deep models \cite{hu2014convolutional,qiu2015convolutional,wan2016deep,pang2016text} are effectively dealing with text snippets, e.g., a pair of sentences, but have difficulty handling extreme short text and long documents in query document matching tasks. On one hand, encoding the query consisting of only few words with complicated deep models usually results in excessive deformation.
On the other hand, it is more likely to introduce ``noise'' and redundant information when dealing with long documents using deep models.

What is more, most existing approaches consider text pieces as sequences of words or word vectors.
However, the semantic structure information of text pieces is not fully utilized, especially when the text length is as long as a document. In the next section, we will introduce our proposed procedures to transform a document into a keyword graph. Such a graph representation proves to be effective at uncovering the underlying attention structure of a long text document such as a news article.
\section{Document as graph}
\label{sec:graph}

\begin{figure}[tb]
\centering
\includegraphics[width=3.3in]{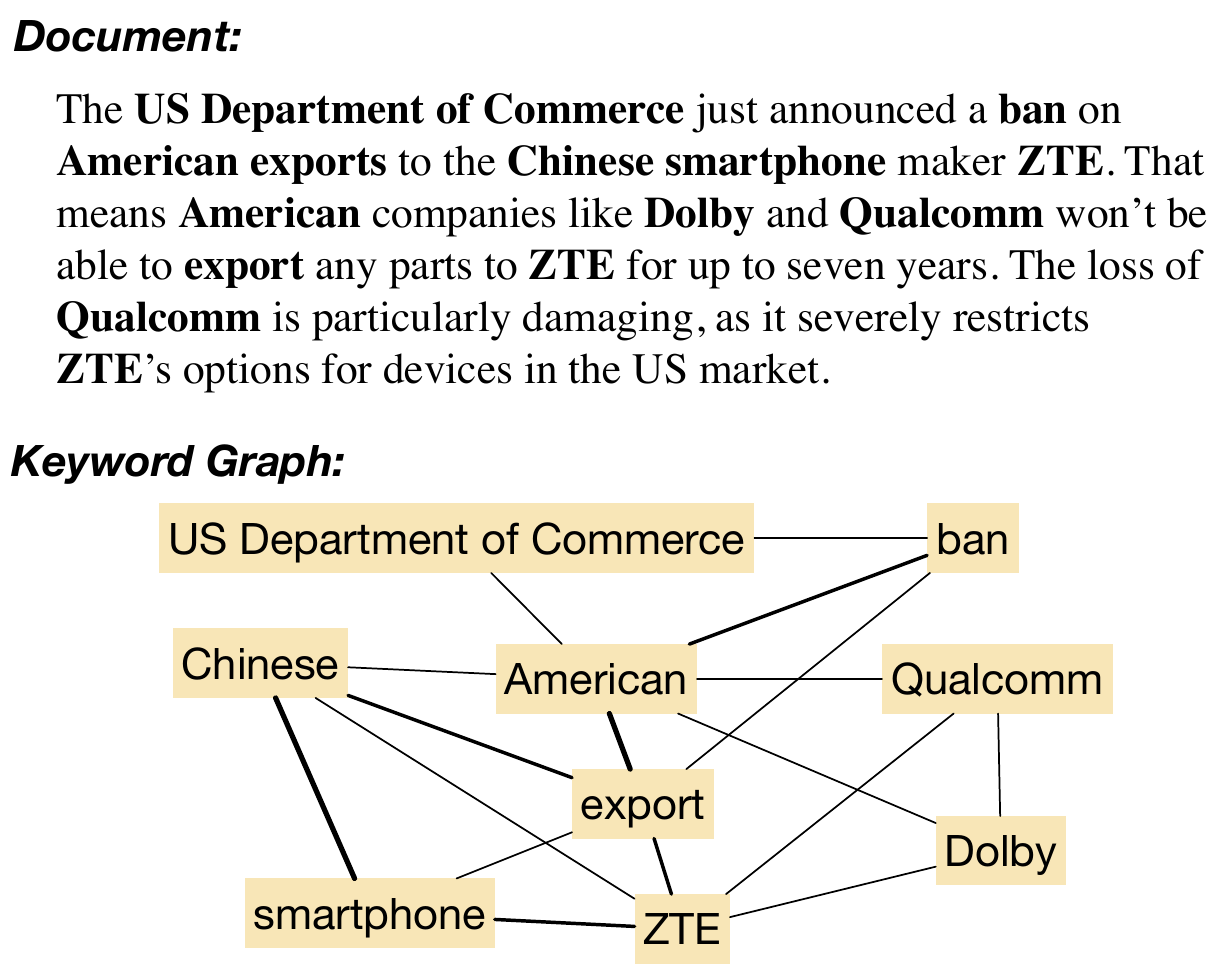}
\caption{An example to show a piece of document and its corresponding Keyword Graph representation.}
\label{fig:CaseStudy}
\vspace{-2mm}
\end{figure}

To address the challenges of the relevance matching problem, we convert the document to a weighted, undirected \textit{keyword graph}. The aim of this graph representation is to model the interaction structure of document keywords, as well as uncovering the term importance of keywords induced by the topological structure of keyword interactions. Compared with linear representation of text pieces, a graphical representation can better capture the rich intrinsic semantic structures in long text objects. Furthermore, it is helpful in overcoming the long-distance dependency problem in NLP, as it breaks the linear organization of words.

We first describe the structure of a document keyword graph before presenting the detailed steps to derive it.
Given an input document $\mathcal{D}$, our objective is to obtain a graph representation $G_D$ of $\mathcal{D}$.
Each vertex in $G_D$ is a keyword in document $\mathcal{D}$.
We connect two vertices by an edge if the word distance of the two keywords in the document is smaller than a threshold (we set the threshold as $20$ in our experiments). The edge weight is proportional to the inverse of the word distance between two keywords.

As a toy example, Fig.~\ref{fig:CaseStudy} illustrates how we convert a document into a keyword graph.
We can extract keywords or key phrases such as \textit{ZTE}, \textit{Qualcomm}, \textit{US Department of Commerce}, \textit{export} and so on from the document using common keyword extraction algorithms \cite{siddiqi2015keyword}.
These keywords represent the topics or concerns in this document.
We then connect the keyword vertices by weighted edges, where the edge weight between a pair of keywords denotes how close the they are related, and the whole topological structure of the keyword graph shows the semantic structure of the document. For example, in Fig.~\ref{fig:CaseStudy}, \textit{export} is highly correlated with \textit{ZTE}, \textit{Chinese}, \textit{American} and so on.
In this way, we have transformed the original document into a graph of different focal points, as well as the interaction topology among them.

\subsection{Keyword Graph Construction}

We now introduce our detailed procedure to restructure a document $\mathcal{D}$ into a desired keyword graph $G_D$ as described above. The whole process consists of three steps: 1) document preprocessing, 2) keyword extraction, and 3) edge construction.

\textbf{Document preprocessing.} The first step is preprocessing the input documents. 
We can utilize off-the-shelf NLP tools such as Stanford CoreNLP~\cite{manning2014stanford} to clean the text and tokenize words.
Then, we extract named entities from the document.
For documents, especially news articles, the named entities are usually critical keywords.

\textbf{Keyword extraction.} The next step is to extract the keywords of documents. As the named entities alone are not enough to cover the main focuses of the document, we therefore apply the keyword extraction algorithm to expand the keyword set.
There are different algorithms for keyword extraction~\cite{siddiqi2015keyword}, such as TF-IDF, TextRank, RAKE and so on.
Since TF-IDF takes the advantages of wide generality and high efficiency, we implemented it in our experiments. 
More specifically, we first calculate the term frequency-inverse document frequency (TF-IDF) value for each token, and choose the top $20$ percentage tokens to expand the set of document keywords.
Even though more sophisticated algorithms may achieve better performance for the keyword extraction, in this paper, we concentrate on the graph modeling of documents and the algorithm of relevance matching. 
After we extract the set of keywords from a document, each keyword will be a vertex in the document's graph.

\textbf{Edge construction.}
Our last step is connecting correlated keywords in the document by weighted edges. For each pair of keyword vertices $v_i$ and $v_j$, we calculate the word distance $d_{ij}$ in the document. Suppose that keyword $v_i$ shows $m$ times in the document and keyword $v_j$ shows $n$ times in the document, with $m \leq n$. 
For the $t_{th}$ keyword $v_i$, we select the $v_j$ that is most close to it, and calculate the word distance $d_{ij}^{t}$. The distance $d_{ij}$ is the mean distance between each $v_i$ and its most nearby $v_j$. Based on the word distance $d_{ij}$, the weight $w_{ij}$ of the edge $e_{ij}$ between $v_i$ and $v_j$ is calculated as
\begin{equation}
\label{eq:weight}
   w_{ij} = g(d_{ij}) = \frac{1}{d_{ij}}= \frac{m}{\sum_{t =1}^{m}d_{ij}^{t}}.
\end{equation}

Now, we have transformed an input document into a weighted undirected graph of keywords. Compared with the original sequential structure, a graph structure organized keywords in terms grants a correlation structure. 
Therefore, the problem of long distance dependency can be alleviated as related keywords are linked by weighted edges. Furthermore, the weighted edges represent the strengths of interactions among these concepts. Together with the topology structure of the whole graph, we can also model the importance of different keyword in the document. A keyword with a lot of edges linking it to other keywords is usually more important than other keywords that only have a few edges. A keyword that has strong connections with other keywords (i.e., the edge weight is large) is typically more important than keywords that only have edges with small weights.

There are also existing works that represent a document as a graph of sentences \cite{balinsky2011document,mihalcea2004textrank}, or construct vertices and edges via more complicated methods, such as linking terms in a document to real world entities or concepts based on some resources. On such example is DBpedia \cite{auer2007dbpedia}, which extracts subject-predicate-object triples from text based on syntactic analysis to build directed edges~\cite{leskovec2004learning}. However, since relevance matching is more focused on the term matching signals between the query and document, we choose to model the correlations between keywords instead of sentences or paragraphs of a document. 
Compared with constructing a keyword graph with complicated mechanisms rooted in the knowledge base or syntactic analysis, which are usually time consuming, we model the structure of keyword correlations by a more efficient procedure described above to make it available for real world industry applications. 
We will see that our keyword graph is both efficient and able to improve the performance of relevance matching tasks when combined with the \textit{Multiresolution Graph Attention Network} model, which will be described in the next section.

\section{Multiresolution Graph Attention Network}
\label{sec:match}

\begin{figure*}
\includegraphics[width=\textwidth]{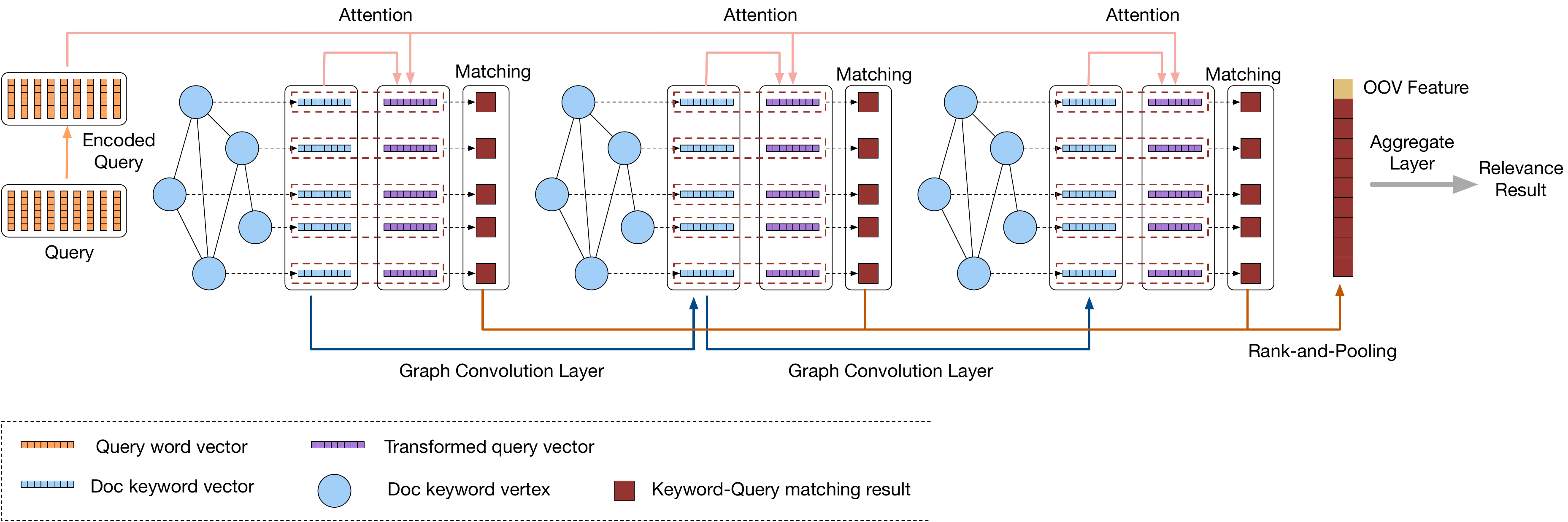}
\caption{An overview of the proposed \emph{Multiresolution Graph Attention Network} (MGAN) for matching a short query and a long text document.}
\label{fig:model}
\vspace{-2mm}
\end{figure*}

In this section, we further exploit the keyword graph representation of documents in Sec. \ref{sec:graph}, and propose a deep relevance model based on multi-layer graph convolutional networks and attention-based matching, namely Multiresolution Graph Attention Network (MGAN), for query document matching.
Fig.~\ref{fig:model} illustrates the overall architecture of our proposed model, which mainly consists of five sequential stages. First, query and vertices in the document graph are embedded with word vectors such as GloVe~\cite{pennington2014glove}. Second, the embedded query and document graph are respectively encoded with convolutional layers. Specifically, for the document graph, graph convolutional layers are implemented to extract the local features of vertices and iteratively revise the encoding vectors. 
Third, a Rank-and-Pooling layer is utilized to sort the vertices in a specific order and unify the graph size. 
Next, we compute the matching scores between query and selected vertices in each graph convolutional layer based on the attention mechanisms.
Finally, these matching scores are concatenated as a match vector and fed into the aggregation layer to get the final relevance matching result.
We will describe each layer in detail as follows.

\subsection{Query Embedding and Encoding}
\label{subsec:queryembed}

The embedding layer turns each token of the query and each keyword of the document into a dense vector. 
Given a query with $d_q$ words, a document graph with $d_g$ vertices and a $d_e$ dimensional pre-trained embedding vectors, we will get a query embedding matrix $Q_{\text{emb}} \in \mathbb{R}^{d_q\times d_e}$ and a graph vertex feature matrix $G_{\text{emb}} \in \mathbb{R}^{d_g\times d_e}$ after the word embedding layer.
In this work, we utilize the pre-trained, 300-dimensional Glove Word Vectors \cite{pennington2014glove} for word embedding in our experiments.
Notice that the out-of-vocabulary (OOV) words, which are not able to be embedded, can still play significant roles in the matching. Especially for a query with only 2 or 3 terms, in this case, each word counts and should not be ignored.
To fully exploit these OOV words, we match them on a term level by calculating how many common OOV words $x_{oov}$ are in the query and document graph. $x_{oov}$ is defined as the OOV feature, and will be concatenated to the final match vector.

It is worth mentioning that we can potentially further improve the performance of our model by combining the character-level embedding with the feature embedding to form the final word representations. 
A character-level embedding of a word (or token) can be obtained by encoding the character sequences with a bi-directional long short-term memory network (BiLSTM) and concatenating the two last hidden states to form the embedding of the token~\cite{hu2017mnemonic}. 
In this way, the meaningful embedding vectors of out-of-vocabulary (OOV) words can also be learned. 

After we embedded the query, we further use a simple 1D convolutional neural network (CNN) as an encoder to produce a refined encoding representation $Q \in \mathbb{R}^{d_q\times d_e}$ of the query, where the $i$-th row in $Q$ is the context vector of token $i$ that incorporates the contextual information in the query.

\subsection{Vertex Encoding based on Graph Convolutional Network}

Unlike the linearly structured query, the document is restructured into a keyword graph. 
After we embedded the vertices by word vectors, we utilize the ability of Graph Convolutional Network (GCN)~\cite{kipf2016semi} to capture the interactions between vertices and get the contextual representation for each vertex.

GCNs generalize traditional CNN from low-dimensional regular grids to high-dimensional irregular graph domains.
Now let us briefly describe the GCN propagation layers in our model, which are used to encode graph vertices with contextual information and revise the vertex vector representation iteratively.
Moreover, we improve the graph convolutional network (GCN) proposed in \cite{kipf2016semi} to better deal with weighted graphs, and learn multiresolution vertex representations through multi-layer graph convolutions. In this way, we can match query and document keywords in different semantic levels and enhance the performance of relevance matching.

\textbf{Graph Convolutional Network for Weighted Graphs.}
Let $\mathcal{G} = (\mathcal{V}, \mathcal{E})$ be an undirected weighted graph consisting of a set of vertices $\mathcal{V}$ with $|\mathcal{V}|=N$ and a set of edges $\mathcal{E}$. To clearly depict the vertex-connection of a graph, the adjacency matrix $A \in \mathbb{R}^{N\times N}$ is introduced, where $A_{ij}$ indicates the weight between vertex $\mathcal{V}_i$ and $\mathcal{V}_j$. The diagonal degree matrix of $A$ is denoted by $D\in \mathbb{R}^{N\times N}$ with $D_{ii}=\sum_{j}A_{ij}$. 

Graph Laplacian, formally defined as $L=D-A\in \mathbb{R}^{N\times N}$, is the fundamental operator in the spectral graph analysis. In addition, there are two normalized versions of the Graph Laplacian, known as Symmetric Laplacian $L_{sys}=I_n-D^{-\frac{1}{2}}AD^{-\frac{1}{2}}$ and Random Walk Laplacian $L_{rw}=I_n-D^{-1}A$ respectively. 
Since the graph $\mathcal{G}$ is undirected and weighted, $L$ is a symmetric positive semidefinite matrix, which can be decomposed to $L=U\Lambda U^{T}$ with a diagonal matrix of eigenvalues $\lambda=diag([\lambda_0, \lambda_1, \cdots , \lambda_{N-1}])$ and a matrix of eigenvectors $U=[u_0, u_1, \cdots , u_{N-1}]$. 

Let us consider the graph convolution in the Fourier domain. As mentioned in \cite{kipf2016semi}, the spectral convolution can be generalized as the Hadamard production of the graph signal and spectral filter in the Fourier domain. Thus, the convolution result $y$ is defined as:
\begin{equation}
\label{eq:sc}
   y=Ug_\theta(\Lambda) U^Tx
\end{equation}
where $x \in \mathbb{R}^N$ is the graph signal with scalar feature for each vertex. Spectral filter $g_\theta(\Lambda)$ is a function of eigenvalues of $L$ parameterized by $\theta \in \mathbb{R}^N$. Note that $\widetilde{x}=U^Tx$ represents the Fourier transform (FT) of the signal $x$, while $U\widetilde{x}$ is the inverse FT. However, the convolution in Eq. \ref{eq:sc} requires explicitly computation of Laplacian eigenvectors, which is not feasible especially for large graphs. To solve this problem, Chebyshev polynomials are implemented to approximate the filter $g_\theta(\Lambda)$ as the K-localized filter $g_\theta^K(\Lambda)$:
\begin{equation}
\label{eq:ca}
    g_\theta(\Lambda)\approx g_\theta^K(\Lambda) = \sum_{k=0}^{K}\theta_kT_k(\widetilde{\Lambda})
\end{equation}
where $\widetilde{\Lambda}=\frac{2}{\lambda_{max}}\Lambda-I_N$ is a diagonal matrix with scaled eigenvalues in the range $[-1,1]$. $\theta=[\theta_0,\theta_1,\cdots,\theta_{K}]$ is a vector of Chebyshev coefficients, and $T_k(\widetilde{\Lambda})$ is the k-th order Chebyshev polynomial evaluated at $\widetilde{\Lambda}$. By the approximation of the filter, Eq. \ref{eq:sc} can be estimated as the K-th localized convolution:
\begin{equation}
\label{eq:ca0}
    y \approx \sum_{k=0}^{K}\theta_kT_k(\widetilde{L})x
\end{equation}
where $\widetilde{L}=\frac{2}{\lambda_{max}}L-I_N$. Recall that Chebyshev polynomials $T_k(\widetilde{L})$ can be derived from a recurrence relation $T_k(\widetilde{L}) = 2\widetilde{L}T_{k-1}(\widetilde{L}) - T_{k-2}(\widetilde{L})$ with $T_0(\widetilde{L})=1$ and $T_1(\widetilde{L})=\widetilde{L}$. In this way, the computation complexity is reduced to $\mathcal{O}(K\left | \mathcal{E} \right |)$. 

Rather than working on all vertices, the K-th localized convolution only focus on the K-hop neighborhoods from the central vertex. Let $K=1$ and $\lambda_{max}=2$, the above model is simplified as:
\begin{equation}
\label{eq:ca1}
     y = \theta_0x+ \theta_1(L-I_N)x
\end{equation}

Properly reduce the number of parameters not only to accelerate computations, but also avoid overfitting in the training process. Unlike parameter settings in \cite{kipf2016semi} with $\theta_0=-\theta_1$, we constrain the parameters to $\theta_0=-\lambda\theta_1$. Denote $\theta_1$ by $\theta$, we have:
\begin{equation}
\label{eq:ca2}
    y = \theta((\lambda+1)I_N-L)x
\end{equation}

Let $X = [x_1, x_2, \cdots ,x_N]^T \in \mathbb{R}^{N\times d_e}$ denotes the vertex feature matrix with each $x_i \in \mathbb{R}^{d_e}$ representing a $d_e$-dimensional feature vector of vertex $\mathcal{V}_i$. When $L=L_{rw}=I_N-D^{-1}A$, the graph convolutional layer can be expressed as:
\begin{equation}
\label{eq:ca3}
    X^{n+1}=\sigma (\widetilde{D}^{-1}( A+\lambda I_N)X^nW^{n}) 
\end{equation}
where $\widetilde{D}_{ii}=\lambda+\sum_{j}A_{ij}$, and $\sigma$ is the active function in each layer such as ReLU.

The parameter $\lambda$ controls the balance between the central vertex and its neighboring vertices. With larger $\lambda$, the central vertex will involve more in the convolutional operation. If $\lambda$ equals to zero, the central vertex will have no contribution to its vertex convolution result.

The convolutional layer of Eq. \ref{eq:ca3} is essentially a generalization of the graph convolutional layer in \cite{kipf2016semi}\cite{zhang2018end} with $\lambda=1$. When Graph Laplacian $L_{sys}=I_n-D^{-\frac{1}{2}}AD^{-\frac{1}{2}}$, the convolution layer becomes the GCN in \cite{kipf2016semi}. However, when $L_{rw}=I_n-D^{-1}A$, it is exactly the same with graph convolutional layer in DGCNN \cite{zhang2018end}. 
Obviously, with the introduced parameter $\lambda$, the graph convolutional layer of Eq. \ref{eq:ca3} can better deal with weighted graph for different scaler of weights. 
For example, if the edge weights are all larger than a hundred, let $\lambda=1$ as it is in GCN~\cite{kipf2016semi} and DGCNN~\cite{zhang2018end}, the central vertex will almost have no influence on its convolution result.

Since the graph convolutional layer can be viewed as a 1-dim Weisfeiler-Lehman algorithm on graphs, for our keyword graph, the convolution process can be interpreted as iteratively revising the representations of vertices based on their neighboring vertices. 
In this way, the contextual information of each vertex in the document is incorporated. With the increasing layers of graph convolution, each vertex will incorporate the information of a broader context (neighbors with a larger distance to it will be considered in the vertex encoding), thus producing a higher level representation of the vertex. Therefore, the multi-layer graph convolution gives multiresolution representations of each vertex.

\subsection{Rank-and-Pooling Layer}
After encoding graph vertices through a multi-layered GCN, we propose a Rank-and-Pooling strategy to sort and select the vertices. 
To be specific, let $X^L = [x_{1}^L, x_{2}^L, \cdots, x_{i}^L, \cdots, x_{d_g}^L]^T$ denotes a $d_g\times d_e$ vertex feature matrix in the last graph convolution layer $L$, where $x_{i}^L=[x_{i1}^L, x_{i2}^L, \cdots, x_{ij}^L, \cdots, x_{id_e}^L]$ is a $d_e$-dimensional feature vector of vertex $\mathcal{V}_i$.
For each dimension $j$ of the vertex features, we normalize it by calculating the softmax over all $d_g$ vertices and then sum up the feature values of all dimensions. That is:
\begin{equation}
\label{eq:ca333}
    T_i = \sum_{j=1}^{d_e}\frac{e^{x_{ij}^L}}{\sum_{i=1}^{d_g}e^{x_{ij}^L}}
\end{equation}

where $T_i$ is the normalized feature sum of vertex $\mathcal{V}_i$. According to the sum $T_i$, $d_g$ vertices are sorted. We then select the top $K$ vertices for further processing.

The Rank-and-Pooling operation is designed for two purposes. First, as there is no order for the vertices in the graph, we use the ranking mechanism to sort the vertices.
Second, the number of keywords $d_g$ (or vertices) varies for different documents. 
We apply the ``max-pooling'' operation to select the top $K$ vertices from each layer to find out the vertices with significant feature values. 
In this way, we can focus on significant keywords for relevance matching, and also control the dimension of the final matching vector.

\subsection{Attention-based Query-Graph Matching}
Based on the above sorted $K$ vertices, we apply an attention matching scheme between the query and selected vertices in each layer.
Given the encoded query matrix $Q \in \mathbb{R}^{d_q\times d_e}$, where $d_e$ is the encoding dimension and $d_q$ is the number of tokens in the query. 
Suppose $\mathbf{v}_i\in \mathbb{R}^{d_e}$ is the $i$-th keyword vertex vector in the graph. For each vertex $\mathcal{V}_i$, we calculate a vertex-aware query representation $\mathbf{q}_i$ as:
\begin{equation}
\label{eq:att}
    \mathbf{q}_i = \text{Attention} (Q, \mathbf{v}_i) = Q \cdot \text{softmax} (Q\cdot\mathbf{v}_i^T),\ \  1 \leq i \leq K. 
\end{equation}
After we get $\mathbf{q}_i$ for each vertex $\mathcal{V}_i$, we then calculate the matching score between query and vertex as
\begin{equation}
\label{eq:att-cos}
    s_{i}^l = \text{CosineSim}(\mathbf{v}_i, \mathbf{q}_i),
\end{equation}
where $s_{i}^l$ denotes the matching score between query and vertex $\mathcal{V}_i$ in the layer $l$.

This layer helps each vertex to focus on the matching signals with a part of the query tokens that are most related to that vertex. If only a small portion of the tokens in the query are correlated to a specific keyword vertex, our attention based query-vertex matching will help to decrease the influence of uncorrelated tokens.

\subsection{Aggregation Layer}
In this layer, we concatenate the matching scores of each vertex in each graph convolution layer, with the OOV feature $x_{oov}$ described above, to form a final matching vector $\mathbf m$ as following:
\begin{equation}
\label{eq:finalvec}
    m = [s_{1}^1, s_{2}^1, \cdots,  s_{K}^1,  \cdots,  s_{k}^l, \cdots, s_{1}^L, s_{2}^L, \cdots, s_{K}^L, x_{oov}],
\end{equation}

where $s_{k}^L$ is the attention matching score between query and $k_{th}$ vertex in the $l_{th}$ layer. Apparently, $m\in \mathbb{R}^{(KL+1)\times 1}$ with $L$ denotes the number of graph convolution layers.

We then feed the concatenated matching vector $\mathbf m$ into a classifier, such as feed forward neural networks, to get the final relevance matching result.

\section{Experiment}
\label{sec:eval}

In this section, our proposed Multiresolution Graph Attention Network is evaluated on two datasets and compared with other existing deep matching models, including both representation-focused deep neural matching models and interaction-focused models.
Then, we further execute an ablation study by removing different parts of our model and evaluating the performance of the model variants. 
The ablation study proves that each module in our model plays a significant role in the task of relevance matching.

\subsection{Description of Tasks and Datasets}
\label{subsec:datasets}

\begin{table}[tb]
  \caption{Description of evaluation datasets.}
  \label{tab:datasets}
  \begin{tabular}{llllll}
    \toprule
    Dataset & Pos & Neg & Train & Dev & Test\\
    \midrule
    Ohsumed & $56976$ & $56976$ & $68370$ & $22789$ & $22793$ \\
    NFCorpus & $64467$ & $35465$ & $59959$ & $19986$ & $19987$ \\
    \bottomrule
  \end{tabular}
  \vspace{-3mm}
\end{table}


In the experiment, we test our model on the following two datasets:
\begin{itemize}
	\item \textbf{Ohsumed dataset for topic document matching  \cite{hersh1994ohsumed}.} The Ohsumed dataset consists of 34394 documents from medical abstracts and are classified into 23 categories of cardiovascular disease groups. The dataset is originally for the document topic classification.
  In our experiment, we generate topic-document pairs from the original dataset, where a positive sample means the topic is the true category of the document. A negative topic-document sample is generated by randomly assigning an incorrect topic to a document.
  The average length of the topic text and documents are $2.6$ and $109.4$.

	\item \textbf{NFCorpus dataset for medical information retrieval.} The NFCorpus dataset is a full-text English retrieval dataset for the task of Medical Information Retrieval. It contains a total of $3,244$ non-technical English queries that harvested from the NutritionFacts.org site, with $169,756$ automatically extracted relevance judgments for $9,964$ medical documents (written in a complex terminology-heavy language), mostly from PubMed \cite{boteva2016full}. We selected a subset of the original dataset which contains $64,467$ samples, as the original dataset is extremely unbalanced. The average length of queries and documents are $3.5$ and $146.7$,  respectively.
\end{itemize}

Table \ref{tab:datasets} shows a detailed breakdown of the datasets used in the evaluation.
For both of the two datasets, we use $60\%$ of samples as a training set to train the model, $20\%$ of samples as a development set to tune the hyper-parameters, and the remaining $20\%$ as a test set.
We train our model by the Adam optimizer with learning rate set to 0.001.
For each model, we carry out training for 5 epochs and then choose the model with the best validation performance for the final evaluation on the test set.

\subsection{Compared Algorithms}
\label{subsec:compare-algs}
We compared our model with the following methods:
\begin{itemize}

\item \textbf{Convolutional Matching Architecture-I (ARC-I)} \cite{hu2014convolutional}: ARC-I is a typical representation-focused deep model, which encodes each piece of text to a vector by CNN and compares the representing vectors with a multilayer perceptron.

\item \textbf{Convolutional Matching Architecture-II (ARC-II)} \cite{hu2014convolutional}: ARC-II is built directly on the local interaction space between two texts, and intends to capture the rich matching patterns at different levels with the $2$-D convolution.

\item \textbf{Deep Structured Semantic Models (DSSM)} \cite{huang2013learning}: DSSM utilizes deep neural networks to map high-dimension sparse features into low-dimensional dense features, and then computes the semantic similarity of the text pair.

\item \textbf{Convolutional Deep Structured Semantic Models (C-DSSM)} \cite{shen2014learning}: C-DSSM learns low-dimensional semantic vectors for input text by CNN. Particularly, DSSM and C-DSSM are designed for Web search. However, they were only evaluated on (query, document title) pairs.


\item \textbf{Multiple Positional Semantic Matching (MV-LSTM)} \cite{wan2016deep}: MV-LSTM matches two texts with multiple positional text representations, and aggregates interactions between different positional representations to give a matching score.


\item \textbf{MatchPyramid} \cite{pang2016text}: MatchPyramid calculates pairwise word matching matrix, and models text matching as image recognition, by taking the matching matrix as an image.



\end{itemize} 

For the above baseline deep matching models, we utilized MatchZoo~\cite{fan2017matchzoo} for evaluation.
For our MGAN model, since the edge weights of the graph is in the range of 0 to 1, we set $\lambda = 1$. 
Besides, considering the average length of documents, $K$ is set to 20 in the Rank-and-Pooling. 
The number of graph convolution layers $L$ is $2$, and the classifier in the aggregation layer is a one-layer feed forward neural network with the hidden size set to $100$.

\subsection{Performance Analysis}
\label{subsec:performance}

\begin{table}[!tb]
  \caption{Accuracy and F1-score results of different algorithms on the Ohsumed dataset.}
  \label{tab:ohsumed}
  \begin{tabular}{c|cc|cc}
    \toprule
    \multirow{2}{*}{Algorithm} & \multicolumn{2}{c}{Dev} & \multicolumn{2}{c}{Test} \\ 
    & Accuracy & F1-score & Accuracy & F1-score \\
    \midrule
    ARC-I & $0.5067$ & $0.6676$ & $0.5068$ & $0.6681$ \\
    ARC-II & $0.5490$ & $0.6759$ & $0.5511$ & $0.6775$ \\
    DSSM & $0.5243$ & $0.4811$ & $0.5138$ & $0.4721$  \\
    C-DSSM & $0.5155$ & $0.5650$ & $0.5112$ & $0.5613$  \\
    MatchPyramid & $0.5597$ & $0.6597$ & $0.5648$ & $0.6625$  \\
    MV-LSTM & $0.5610$ & $0.4559$ & $0.5555$ & $0.4481$  \\
    MGAN & $\mathbf{0.8040}$ & $\mathbf{0.8090}$ & $\mathbf{0.8075}$ & $\mathbf{0.8118}$ \\
    \bottomrule
  \end{tabular}
  \vspace{-3mm}
\end{table}

\begin{table}[!tb]
  \caption{Accuracy and F1-score results of different algorithms on the NFCorpus dataset.}
  \label{tab:nfcorpus}
  \begin{tabular}{c|cc|cc}
    \toprule
    \multirow{2}{*}{Algorithm} & \multicolumn{2}{c}{Dev} & \multicolumn{2}{c}{Test} \\ 
    & Accuracy & F1-score & Accuracy & F1-score \\
    \midrule
    ARC-I & $0.5067$ & $0.6676$ & $0.7969$ & $0.8548$ \\
    ARC-II & $0.5490$ & $0.6759$ & $0.7576$ & $0.8361$ \\
    DSSM & $0.5243$ & $0.4811$ & $0.6336$ & $0.7646$  \\
    C-DSSM & $0.5155$ & $0.5650$ & $0.6259$ & $0.7590$  \\
    MatchPyramid & $0.5597$ & $0.6597$ & $0.6408$ & $0.7811$  \\
    MV-LSTM & $0.5610$ & $0.4559$ & $0.6683$ & $0.7564$  \\
    MGAN & $\mathbf{0.9425}$ & $\mathbf{0.9553}$ & $\mathbf{0.9407}$ & $\mathbf{0.9535}$ \\
    \bottomrule
  \end{tabular}
  \vspace{-3mm}
\end{table}

Table~\ref{tab:ohsumed} and Table~\ref{tab:nfcorpus} compares our model with existing deep matching models on the Ohsumed dataset and the NFCorpus dataset, in terms of classification accuracy and F1 score.
Results demonstrate that our Multiresolution Graph Attention Network achieves the best classification accuracy and F1 score on both two datasets.
This can be attributed to multiple characteristics of our model.
First, the input to our neural network model is the keyword graph representation of documents, rather than the original sequential word representation. Based on it, we characterize the interaction patterns between different keywords of the document. This helps to incorporate the semantic structure information of a long document into our model, and alleviates the problem of long-distance dependency (as correlated words are connected by edges directly).
Our model solves the problem of matching query and document in a ``divide-and-conquer'' manner to cope with the long length of documents: it matches the query with each keyword of the document to get matching signals, and aggregate all the matching signals by utilizing the correlations between keywords to give an overall relevance matching result.
Second, our model learns a multiresolution encoding representation for each keyword vertex via a multi-layer Graph Convolutional Network. In each graph convolution layer, the representations of vertices are revised by taking their neighboring vertices into account. In this way, the context information of the keywords in the document is encoded into the high-level vertex representations.
Third, for each vertex in each graph convolution layer, we learn a vertex-specific query representation through attention mechanism to match the query with each vertex. This operation helps the vertices to focus on the query information that is related to it.
Finally, our rank-and-pooling operation unifies the number of vertices for different documents, and selects the most important matching signals in each layer to get the final matching result.

Table~\ref{tab:ohsumed} and Table~\ref{tab:nfcorpus} indicate that the baseline deep text matching models lead to bad performance in query document relevance matching tasks. 
The main reasons are the following.
First, existing deep text matching models are more suitable for the task of semantic matching, where the main concerns in such tasks are the compositional meanings of text pieces and the global matching between them. In our case, matching query and document is the problem of relevance matching. This problem emphasizes more on the exact matching signals between query keywords and documents. Both the importance of different query keywords and the topic structure of documents are critical to relevance matching, and we need to take them into account.
Second, existing deep text matching models can hardly capture meaningful semantic relations between a short query and a long document. When the document is long, it may covers multiple topics, and the query may match only a part of the document. In this case, it is hard to get an appropriate context vector representation for relevance matching, and the part of document that is not related with the query will overwhelm the match signals of the related part. For interaction-focused models, most of the interactions between words in the query and the document will be meaningless, therefore it is not easy to extract useful interaction features for further matching steps. Our model effectively solves the above challenges by representing documents as keyword graphs, and utilize the semantic structure of long documents through Graph Convolution Network for relevance matching.

We also tried to represent the query and document by TF-IDF vector, and then calculate the cosine similarity to estimate the relevance level between them. We found that the performance given by such Bag-of-Word models are quite bad (the accuracy is around 0.38 and F1 score is smaller than 0.1) because of the extremely sparse vector of the query. This proves the necessity of representing words by word vectors, and incorporating document structural information by graph convolution.

In overall, the experimental results demonstrate the superior applicability and generalizability of our proposed model.

\subsection{Impact of Different Modules and Parameters}

\begin{table}[tb]
  \caption{Accuracy and F1-score results of MGAN and its variants on the Ohsumed dataset.}
  \label{tab:ablation-oh}
  \begin{tabular}{l|cc|cc}
    \toprule
    \multirow{2}{*}{Algorithm} & \multicolumn{2}{c}{Dev} & \multicolumn{2}{c}{Test} \\ 
    & Accuracy & F1-score & Accuracy & F1-score \\
    \midrule
    No GCN & $0.6837$ & $0.6819$ & $0.6850$ & $0.6810$ \\
    No Attention & $0.7908$ & $0.7882$ & $0.7893$ & $0.7865$ \\
    No Query Encoding & $0.7859$ & $0.7900$ & $0.7927$ & $0.79576$  \\
    Pooling Size $K=5$ & $0.7602$ & $0.7453$ & $0.7642$ & $0.7484$  \\
    MGAN & $\mathbf{0.8040}$ & $\mathbf{0.8090}$ & $\mathbf{0.8075}$ & $\mathbf{0.8118}$ \\
    \bottomrule
  \end{tabular}
  \vspace{-4mm}
\end{table}

\begin{table}[tb]
  \caption{Accuracy and F1-score results of MGAN and its variants on the NFCorpus dataset.}
  \label{tab:ablation-nf}
  \begin{tabular}{l|cc|cc}
    \toprule
    \multirow{2}{*}{Algorithm} & \multicolumn{2}{c}{Dev} & \multicolumn{2}{c}{Test} \\ 
    & Accuracy & F1-score & Accuracy & F1-score \\
    \midrule
    No GCN & $0.8767$ & $0.9053$ & $0.8757$ & $0.9039$ \\
    No Attention & $\mathbf{0.9432}$ & $\mathbf{0.9558}$ & $\mathbf{0.9433}$ & $\mathbf{0.9556}$ \\
    No Query Encoding & $0.8616$ & $0.8929$ & $0.8629$ & $0.8930$  \\
    Pooling Size $K=5$ & $0.9381$ & $0.9520$ & $0.9381$ & $0.9517$  \\
    MGAN & $0.9425$ & $0.9553$ & $0.9407$ & $0.9535$ \\
    \bottomrule
  \end{tabular}
  \vspace{-4mm}
\end{table}

We also tested several model variants for ablation study. For each model variant, we remove one module from the complete Multiresolution Graph Attention Network model, and compare its performance with our complete model on the two datasets to evaluate the impact of the removed component.

Table~\ref{tab:ablation-oh} and ~\ref{tab:ablation-nf} show the performance of all evaluated models for ablation study. Specifically, we evaluated the following models:
\begin{itemize}
  \item \textbf{MGAN}. This is our original proposed model.
  \item \textbf{MGAN (no GCN)}. This is a variant model that removes the graph convolutional layers in the MGAN. In other words, we represent each vertex by the word vector, and match each keyword with all query terms.
  \item \textbf{MGAN (no attention)}. This variant model deletes the attention mechanism in the MGAN. In this model, we add a max-pooling layer over the encoded query words to get the hidden vector representation of the query, and use it to match with each vertex.
  \item \textbf{MGAN (less keywords)}. In this model, we reduce the number of selected keywords by setting $K=5$ instead of $20$.
  \item \textbf{MGAN (no query encoding)}. In this model, we remove the 1D CNN encoder for query, and directly use the word vectors to represent each query token.
\end{itemize}


\textbf{Impact of graph convolution layers.}
Compare our model with the version that do not contain any graph convolution layers, the performance is worse on both datasets when we remove graph convolution from our model. The reason is that the representation of each vertex will be local and does not contain any context information of its neighboring vertices. Therefore, the topological structure of keyword interactions in the document are totally ignored. In our model with graph convolution layers, in each layer, we lean an adaptive context vector for each vertex. It incorporates the semantic meaning of its neighboring keywords based on their vector representations and edge weights. The multi-layer graph convolution leads to a multiresolution semantic representation of keywords in the document, as in a higher layer, the representation of a vertex covers the information of vertices in a broader range.

\textbf{Impact of query encoding.}
Compare our model with the version that do not perform query encoding. When the query tokens are only represented by the original word vectors and not refined by any encoders to incorporate the contextual information, the performance becomes worse. For example, if the main focus of the query is a key phrase that contains multiple tokens, the CNN encoder can help to combine the semantic information in tokens to represent the key phrase, while the original sequence of word vectors can hardly capture the compositional meaning.

\textbf{Impact of query-vertex attention.}
Compare our model with the version that do not implement query-vertex attention. In this case, our model gets better performance on the Ohsumed dataset and comparable performance on the NFCorpus dataset. Our model use the attention mechanism to learn a vertex-aware query encoding for each vertex. Thus, each vertex will focus on the matching signals with a subset of the query tokens. 
In comparison, when we remove the attention mechanism from our model, each vertex will match with the same encoding vector of the query. 
Given a specific vertex, the unrelated tokens in the query make the matching signal between a query and a keyword less important.
However, when tokens in the query have similar meaning, the attention mechanism won't have significant impact on the performance of our model.

\textbf{Impact of the number of selected keywords in the Rank-and-Pooling.}
In the Rank-and-Pooling operation, we need to set a parameter $K$ and choose the matching results between the query and the top $K$ vertices for each graph convolution layer. 
We tested $K=5$ and $K=20$ respectively, and the performance is better when $K=20$. 
That is reasonable, as $K=20$, our keyword graphs of documents retain more information of the original documents. 
If the value of $K$ is small, keywords related to the query are more likely to be removed. 
However, if the value of $K$ is too large, the unimportant words in the document will become noise to the matching model thus leading to bad performance. 
Furthermore, we should also take the time complexity of the model into account.
More vertices selected in each layer, the more time we need for computation.



\begin{figure}[t]
                        \centering
                        \subfigure[Compare Accuracies]{
                \includegraphics[width=1.6in]{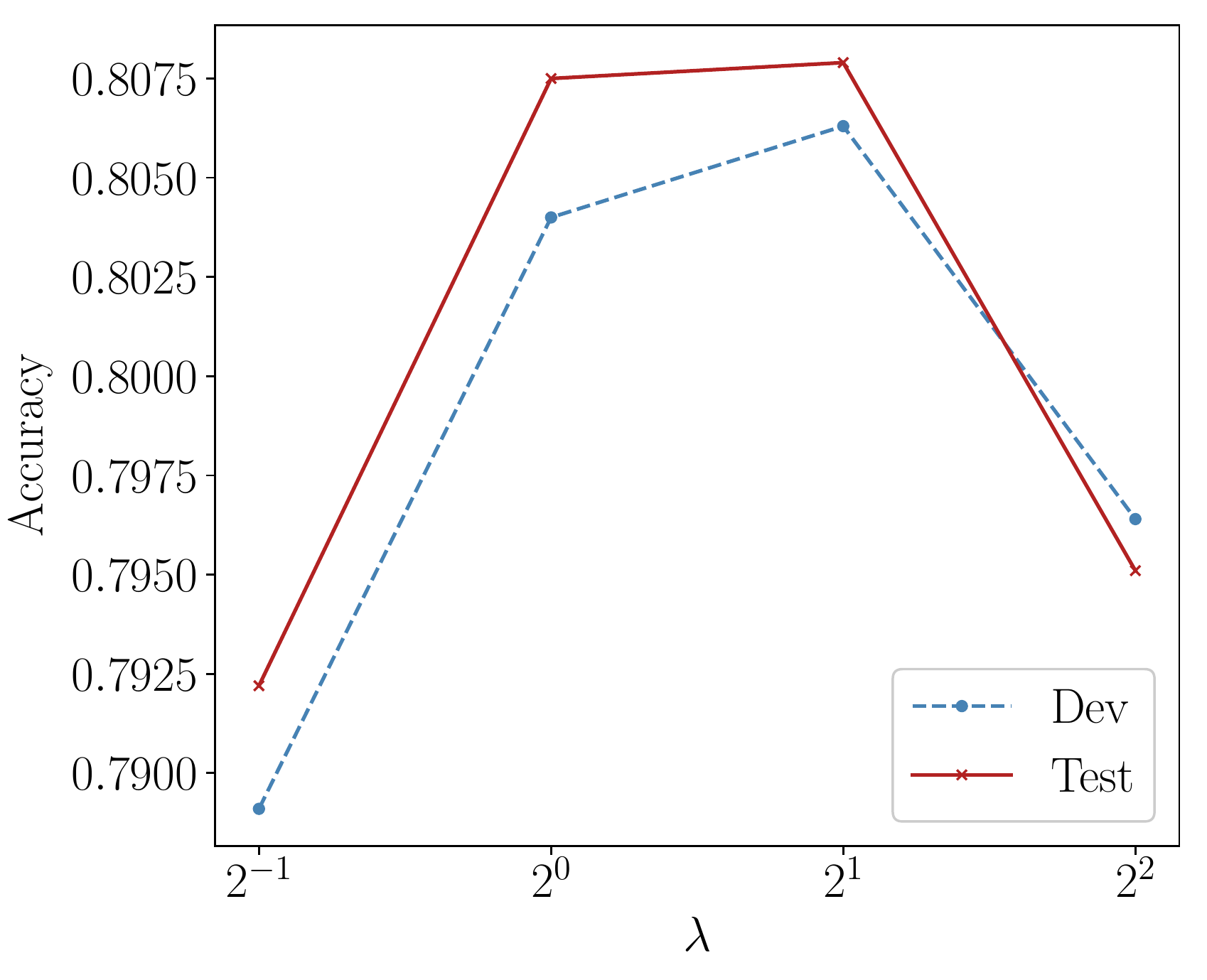}
                                \label{fig:barplotA}
                        }
                \hspace{-2mm}
                        \subfigure[Compare F1 Scores]{
                \includegraphics[width=1.56in]{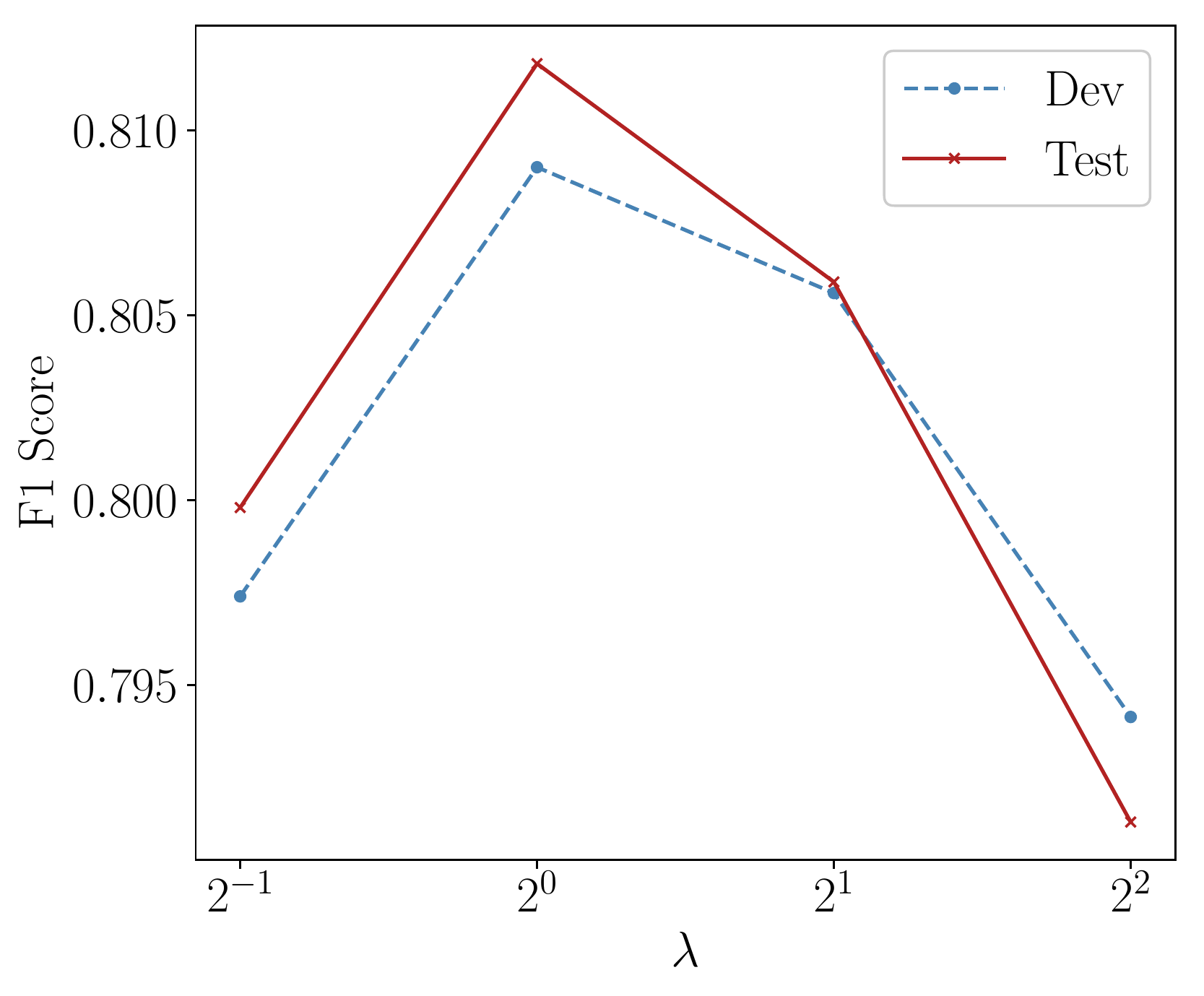}
                                \label{fig:barplotB}
                        }
                        \vspace{-2mm}
                \caption{Compare the accuracies and F1 scores given by different $\lambda$ on the Ohsumed dataset.}
                \label{fig:lambda}
\vspace{-5mm}
\end{figure}

\textbf{Impact of parameter $\lambda$}.
We tested the performance of our MGAN model on the Ohsumed dataset with different values of $\lambda$. Fig. \ref{fig:lambda} shows the comparison result in terms of accuracy and F1 score. 
It shows that the performance of our model achieve the best when $\lambda$ is set to be around $1$. 
If $\lambda$ is too small or too large, the accuracy and F1 score will decrease. 
The reason is that the value of $\lambda$ shall be around the same scale with the edge weights in the keyword graph. In our experiments, the edges weights are within the range of $0$ to $1$. 
Large $\lambda$ means that we focus more on each vertex's own information and incorporate little contextual information into it by graph convolution. 
In contrast, a small value of $\lambda$ makes the graph convolution emphasize on incorporating the contextual information of vertex's neighboring vertices, but the vertex's own information plays a less important role. 
Therefore, $\lambda$ is significant to the weighted graphs and should set to an appropriate scale.

\section{Related Work}
\label{sec:related}

There are mainly two research lines that are highly related to our work: Document Graph Representation and Text Matching.

\subsection{Document Graph Representation}
\label{sec:related-docgraph}
Various of graph representations have been proposed for document modeling. Based on the different types of graph nodes, a majority of existing works can be generalized into four categories: word graph, text graph, concept graph, and hybrid graph.

For word graphs, the graph nodes represent different non-stop words in a document. \cite{leskovec2004learning} extracts subject-predicate-object triples from text based on syntactic analysis, and merge them to form a directed graph. 
\cite{rousseau2013graph,rousseau2015text} represent a document as graph-of-word, where nodes represent unique terms and directed edges represent co-occurrences between the terms within a fixed-size sliding window. \cite{wang2011representing} connect terms with syntactic dependencies. 

Text graphs use sentences, paragraphs or documents as vertices, and establish edges by word co-occurrence, location or text similarities.  \cite{balinsky2011document,mihalcea2004textrank,erkan2004lexrank} connect sentences if they near to each other, share at least one common keyword, or the sentence similarity is above a threshold. \cite{page1999pagerank} connects web documents by hyperlinks. 
\cite{putra2017evaluating} constructs directed graphs of sentences for text coherence evaluation. 

Concept graphs connect terms in a document to real world entities or concepts based on resources such as DBpedia \cite{auer2007dbpedia}, WordNet \cite{miller1995wordnet}, VerbNet \cite{schuler2005verbnet} and so forth. \cite{hensman2004construction} identifies the semantic roles in a sentence with WordNet and VerbNet, and combines these semantic roles with a set of syntactic rules to construct a concept graph.

Hybrid graphs contains multiple types of vertices and edges. \cite{rink2010learning} uses sentences as vertices and encodes lexical, syntactic, and semantic relations in edges. \cite{jiang2010text} extract tokens, syntactic structure nodes, semantic nodes and so on from each sentence, and link them by different types of edges. 


\subsection{Text Matching}
\label{sec:related-textmatching}

The most straight forward method for text matching in information retrieval is lexical matching \cite{berry1995using}, which matches terms in the query with those in the document. However, term level matching suffers from synonymy as well as polysemy. 
Instead of directly matching the words, bag-of-words (BOW) model matches text based on statistics. 
For BOW model, text is vectorized with TF-IDF to evaluate the co-occurrence of words. We then calculate the distance or similarity between vectors with euclidean distance, cosine correlation, etc. 
Besides, another metric Okapi BM25 \cite{robertson2009probabilistic} based on the probabilistic model is also widely implemented in the industry.
However, these models are based on the assumption that words in the text are independent, disregarding the word order and semantic meaning of each word.
Topic models such as latent semantic indexing (LSI) \cite{rosario2000latent}, is designed to explore the second-order co-occurrence in the text with singular value decomposition (SVD).
Feature-based models, like IRGAN \cite{wang2017irgan}, are effective. However, they rely on hundreds of handcrafted features, which are time-consuming, incomplete and over-specified. 

Considering both word semantics and word sequences, deep matching models have seen great success in recent years.
Deep matching models can be divided into two categories depending on the models' architecture: representation-focused model and interaction-focused model.
Representation-focused deep matching models usually transform the word embedding sequences of text pairs into context representation vectors through a neural network encoder, followed by a fully connected network or score function which gives the matching result based on the context vectors. 
Such models include ARC-I \cite{hu2014convolutional}, DSSM \cite{huang2013learning}, C-DSSM \cite{shen2014learning} and so on. 
Interaction-focused models build local interactions between words or phrases to extract the matching features. Then aggregate the matching features to give a matching result.
Models such as ARC-II \cite{hu2014convolutional}, DeepMatch \cite{lu2013deep} and MatchPyramid \cite{pang2016text} are all interaction-focused. 
However, the intrinsic structural properties of long text documents are not fully utilized by these neural models. Our model combines the graphical representation of documents and Graph Convolutional Network to incorporate the structural information for relevance matching.

\section{Conclusions}
\label{sec:conclude}

In this paper, we point out the key role of semantic structures of documents in the task of relevance matching between \textit{short-long} text pairs, and show that most existing approaches cannot achieve satisfactory performance for this task. 
We propose to model a long document as a weighted undirected graph of keywords, with each vertex representing a keyword in the document, and edges indicating their interaction levels. 
Based on the graph representation of documents, we further propose the \textit{Multiresolution Graph Attention Network} (MGAN), a novel deep neural network architecture, which learns multi-layer representations for keyword vertices through a Graph Convolutional Network. It models the local interactions between query words and each document keyword based on attention mechanism, and combines the multiresolution matching between query and keywords on different graph convolution layers with a \textit{rank-and-pooling} procedure to give the final relevance estimation result.
We apply our techniques to the task of relevance matching based on the Ohsumed dataset and the NFCorpus dataset. The simulation results show that the proposed approach can achieve significant improvement for relevance matching in terms of accuracy and F1 score, compared with multiple existing approaches.

\bibliographystyle{ACM-Reference-Format}
\bibliography{main}

\end{document}